\pdfoutput=1

\documentclass[11pt]{article}

\usepackage[final]{acl}

\usepackage{times}
\usepackage{latexsym}

\usepackage[T1]{fontenc}

\usepackage[utf8]{inputenc}
\usepackage{url}
\usepackage{microtype}
 \usepackage{amsmath}
\usepackage{inconsolata}
\usepackage[utf8]{inputenc}
\DeclareUnicodeCharacter{202A}{}

\usepackage{graphicx}
\title{Au-M-ol: A Unified Model for Medical Audio and Language Understanding}
\author{Meizhu Liu\textsuperscript{1}\thanks{These authors contributed equally.} \quad Nistha Mitra\textsuperscript{2}\footnotemark[1]\quad  Paul Li\textsuperscript{1} \quad Amine Abdaoui\textsuperscript{1} \quad Adam Ledyard\textsuperscript{1} \quad Tao Sheng\textsuperscript{1}
 \\[0.5em]
  \textsuperscript{1}Oracle AI Science \quad
  \textsuperscript{2}Neuramill}

\date{\today}

\begin{document}
\maketitle

\begin{abstract}

In this work, we present Au-M-ol, a novel multimodal architecture that extends Large Language Models (LLMs) with audio processing. It is designed to improve performance on clinically relevant tasks such as Automatic Speech Recognition (ASR). Au-M-ol has three main components: (1) an audio encoder that extracts rich acoustic features from medical speech, (2) an adaptation layer that maps audio features into the LLM input space, and (3) a pretrained LLM that performs transcription and clinical language understanding. This design allows the model to interpret spoken medical content directly, improving both accuracy and robustness.  In experiments, Au-M-ol reduces Word Error Rate (WER) by 56\% compared to state-of-the-art baselines on medical transcription tasks. The model also performs well in challenging conditions, including noisy environments, domain-specific terminology, and speaker variability. These results suggest that Au-M-ol is a strong candidate for real-world clinical applications, where reliable and context-aware audio understanding is essential. 
\end{abstract}

\section{Introduction}
The global healthcare system has faced mounting pressure in recent years, with the COVID-19 pandemic intensifying deep-rooted structural challenges \cite{Smyrnakis2021, liu2016}. Chronic issues such as overcrowded hospitals, lengthy patient waitlists, and critical shortages of healthcare professionals have worsened, leading to widespread burnout and increasing attrition among clinicians \cite{Smyrnakis2021, Zhang2020}. As healthcare systems struggle to meet rising demand with limited resources, the need for scalable, intelligent solutions to enhance operational efficiency and support clinical workflows has become more urgent than ever.
Artificial intelligence (AI), particularly the latest large language models (LLMs), offers transformative potential for healthcare delivery. These models can streamline documentation, improve clinical decision-making, and enable new forms of interaction with medical data. However, current LLMs and multimodal models remain limited in their applicability to complex clinical environments—particularly when audio is involved.

Models such as Whisper \cite{Radford2022}, Audio Qwen \cite{Chu2023}, MaLa-ASR \cite{yang2024malaasr}, and Macaw-LLM \cite{lyu2023macaw} have demonstrated strong performance in general-purpose audio transcription. Building on this foundation, a growing body of research has investigated joint modeling of audio and text. Approaches like Speech-LLaMA \cite{speechllama2023}, mSLAM \cite{mslam2022}, SpeechUT \cite{speechut2022},  SSR \cite{ssr2024} and some others \cite{Naderi2024, Karner} leverage cross-modal alignment techniques or adapter modules to fuse acoustic and linguistic representations. Meanwhile, architectures such as the Perceiver \cite{perceiver2021, liu2013, Du20} and Dual-decoder Transformer \cite{dualdecoder2020} aim to efficiently integrate heterogeneous modalities at scale. More recent models, including the Cascaded Cross-Modal Transformer \cite{cascaded2024}, show promise in higher-level tasks such as emotion and intent recognition from audio-text pairs.

\begin{figure*}[th]
  \includegraphics[width=1\linewidth]{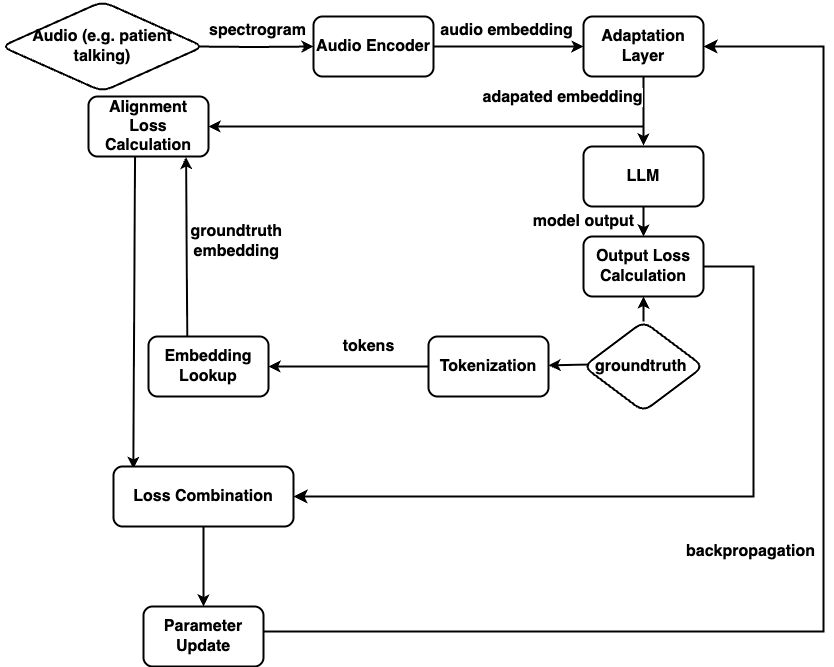}
  \caption{The overview of the Au-M-ol model architecture. }
  \label{fig:diagram}
\end{figure*}

Despite these advances, such models are typically trained and evaluated on general-domain data and often fall short in the medical domain, where language tends to be more complex and highly specialized, and clinical accuracy is paramount. Whisper, while highly capable at speech-to-text transcription, does not capture essential clinical audio biomarkers—such as intonation, breathing patterns, or acoustic irregularities—that are critical for diagnosing conditions like Parkinson’s disease or respiratory disorders. Similarly, Audio Qwen \cite{Chu2023} incorporates multimodal signals but lacks the semantic grounding necessary for interpreting domain-specific intent. Models like SSR and SpeechUT focus on modality alignment but have not been applied to high-stakes clinical applications where interpretability and robustness are essential.

Med-PaLM and Med-PaLM 2 \cite{Google2022} are healthcare-focused models that achieve strong performance in medical question answering, reasoning, and knowledge retrieval. Despite their textual reasoning capabilities, these models are unable to process or transcribe audio, limiting their applicability in clinical scenarios that require automatic speech recognition (ASR).

Some recent approaches, such as \emph{The Sound of Healthcare} \cite{Adedeji24}, combine ASR with LLMs in a sequential pipeline to improve transcription quality for medical audio. While effective, this two-stage architecture introduces several limitations:

\begin{itemize} 
    \item \textbf{High computational cost:} Running separate ASR and LLM models increases inference time and hardware requirements.
    \item \textbf{Scalability challenges:} Maintaining two independent models complicates deployment across large healthcare systems.
    \item \textbf{Fragmented processing:} Passing intermediate outputs between models introduces potential failure points and data inefficiencies.
\end{itemize}

These limitations can hinder the responsiveness and scalability of such systems, restricting their adoption in time-sensitive, real-world healthcare applications. 

More recently, \emph{Whispering-LLaMA} \cite{radhakrishnan2023whispering} integrates audio embeddings from Whisper directly into LLaMA via cross-attention adapters. This allows LLaMA to reason over both audio and text simultaneously, resulting in more accurate transcriptions, particularly in challenging scenarios. However, because both the Whisper decoder and the LLaMA decoder are involved, two transformer decoding passes with cross-modal attention are required, which can slow down real-time applications.

To address these limitations, we propose Au-M-ol, a novel multimodal architecture that unifies audio and text processing into a single, efficient encoder-decoder framework. Our model integrates an audio encoder with a decoder-based LLM via an adaptation layer that aligns acoustic features with language embeddings. This tight integration allows the model to perform transcription and understanding in a unified manner, significantly reducing latency and computational overhead.
Unlike traditional cascaded systems, Au-M-ol is lightweight, easier to fine-tune, and capable of real-time inference. It outperforms state-of-the-art baselines in medical transcription tasks, achieving significant transcription accuracy in medical audios. 

The remainder of the paper is organized as follows. Section \ref{sec:proposedmodel} presents the details of the proposed model. Section \ref{sec:experiments} describes the experimental setup, including the datasets, training procedures, and evaluation results. Section \ref{sec:conclusions} provides concluding remarks, Section \ref{sec:futurework} talks about possible future directions and Section \ref{sec:limitations} discusses the limitations of our work.

\section{Proposed Model}
\label{sec:proposedmodel}


The core of the model is a lightweight, robust encoder-decoder transformer architecture that extracts health-related information from speech and processes it alongside text using a decoder-based language model. This unified design eliminates the need for separate systems for transcription, documentation, and diagnostics, significantly simplifying workflows and reducing complexity.

\subsection{Model Architecture}

Au-M-ol is a unified encoder-decoder architecture that integrates speech and text understanding within a single model. The model consists of a Whisper encoder, a lightweight adaptation layer, and a decoder-only language model (LLaMA-3.1-8B).
Unlike prior work that adopts a sequential ASR-to-LLM pipeline \cite{Adedeji24, Google2022}, our model jointly processes audio (e.g., prosody, speech patterns) and text (e.g., clinical notes) in a single end-to-end framework. This integration removes the need for separate ASR and language modeling components, reducing complexity and improving efficiency.
The overall architecture is shown in Figure~\ref{fig:diagram}. We describe each component in detail below.

\subsection{Audio Encoder}
 


We employ a Whisper-based audio encoder (Whisper-Large-V2) \cite{Radford2022} to transform raw audio waveforms into high-level feature representations. The encoder is composed of a stack of Transformer blocks, each consisting of multi-head self-attention, feed-forward layers, layer normalization, and residual connections.

Prior to encoding, audio is preprocessed through several steps. First, the waveform is resampled to 16 kHz. It is then converted into a log-Mel spectrogram using a short-time Fourier transform (STFT) with a window size of 25 ms and a hop (stride) of 10 ms. The spectrogram is mapped into 80 Mel frequency bins, resulting in a perceptually-informed representation aligned with human auditory perception. Each input covers a fixed duration of 30 seconds, yielding a spectrogram tensor of shape (80, 3000), where 3000 corresponds to the number of frames.

The spectrograms are normalized per frequency bin, and positional encodings are added to preserve the temporal structure of the input. These are learned sinusoidal embeddings, enabling the Transformer layers to capture sequence order and long-range dependencies across time. The resulting representation is then passed through the Whisper encoder to produce contextualized audio features, which will be input to the adaptation layer. 


\subsection{Adaptation Layer}

The adaptation layer acts as a bridging component between the audio encoder and the language model, performing both dimensionality reduction and feature transformation to map acoustic representations into the LLM embedding space. It subsamples the high-dimensional output of the Whisper encoder while preserving essential temporal and spectral information. To further align modalities, cross-modal attention mechanisms within the adaptation layer establish explicit correspondences between acoustic features and linguistic representations, enabling the construction of a shared embedding space.
The adaptation module consists of the following operations:

\begin{itemize}
\item Fully connected linear layer: Projects the audio encoder outputs into a hidden-dimensional space (e.g., 2048). The dimension size can be optimized based on the training dataset.
\item ReLU activation: Introduces non-linearity, allowing the model to learn complex mappings.
\item Fully connected linear layer: Transforms the intermediate representation to match the dimensionality of the LLM's embedding space (e.g. dimension 4096).
\item Layer normalization: Normalizes feature values across each sample, improving training stability and convergence.
\end{itemize}
This transformation ensures that the encoded audio features are compatible with the language model, enabling unified multimodal processing within a shared vector space.




\subsection{LLM Decoder}

The language model in our framework is a decoder-only architecture (e.g., LLaMA-3.1-8B). The audio embeddings produced by the adaptation layer are aligned with the LLM's text embeddings, enabling the model to jointly reason over both auditory and textual modalities. This multimodal integration allows the system to capture richer clinical cues—such as intent, condition markers, and emotional states—leading to a more comprehensive understanding of a patient's health status.

To promote a shared semantic space between modalities, we compute an alignment loss between the text embeddings and the adapted audio features. This loss encourages the model to learn meaningful cross-modal representations, improving its ability to integrate and interpret multimodal clinical inputs.

During generation, the decoder receives a sequence formed by concatenating the adapted audio embeddings with a tokenized textual prompt. The model attends to both sources of information to produce multimodal outputs. These outputs are then compared to reference ground-truth responses using a supervised output loss, enabling the model to learn from both semantic alignment and task-specific supervision.

\subsection{Loss Function}

Our loss function jointly optimizes both transcription accuracy and audio-text alignment through two components: \textit{output loss} $\mathcal{L}_{\text{output}}$ and \textit{alignment loss} $\mathcal{L}_{\text{alignment}}$. The output loss is computed using the cross-entropy loss function, which measures the divergence between the model's predicted output and the ground truth labels. This encourages the model to generate accurate transcriptions.

The alignment loss aligns the projected audio features $e_t$ with the expected decoder input $e_u$ embeddings. It is a combination of L1 loss and cosine similarity loss 

\begin{equation}
  \label{eq:alignloss}
\mathcal{L}_{\text{alignment}} = \mathcal{L}_{1}(e_t, e_u) + 1 - cos(e_t, e_u)
\end{equation}
Here the $\mathcal{L}_{1}$ loss captures the absolute differences between vectors and is robust to outliers, and the $cos(e_t, e_u)$ similarity  measures the cosine of the angle between two vectors in the embedding space, reflecting their directional similarity. The output loss and the alignment loss are combined using a weighting parameter $\alpha$ to compute the final total loss:
\begin{equation}
  \label{eq:loss}
\mathcal{L}_{\text{total}} = \mathcal{L}_{\text{output}} + \alpha \cdot \mathcal{L}_{\text{alignment}},
\end{equation}
where $\alpha$ is a hyperparameter tuned based on the training data. In our experiments, setting $\alpha=1$ yielded effective results. This formulation enables the model to jointly optimize for transcription accuracy and cross-modal representation alignment.

\subsection{Model Advantages}


By leveraging a single, unified model, our approach addresses several longstanding challenges in healthcare AI:

\begin{itemize}
    \item High Computational Cost: Integrating the audio encoder and language model into a single architecture reduces resource consumption, eliminating the inefficiencies of running separate ASR and LLM systems.
\item Latency: The end-to-end design removes the need for sequential post-processing, enabling faster inference and improved real-time responsiveness.
\item Fragmented Processing: By jointly processing audio and text within the same model, we eliminate inter-system data transfer and reduce potential points of failure.
\item Adaptability: The model supports flexible prompt engineering, allowing clinicians to tailor outputs to specific needs—e.g., summarizing patient histories or generating diagnostic narratives—without retraining. Furthermore, it supports audio-based Retrieval-Augmented Generation (Audio-RAG), enabling real-time access to relevant clinical knowledge.
\end{itemize}

\section{Experimental Settings}
\label{sec:experiments}
We trained and evaluated the model on various datasets, using our customized training pipeline. In this section, we describe the datasets employed, detail the training methodology, and present the experimental results, including comparisons with state-of-the-art models.

\subsection{Datasets Used}


We utilized a diverse collection of datasets for both model training and evaluation. Publicly available datasets include FLEURS (English) \cite{FLEURS} and LibriSpeech \cite{Librispeech}.  In addition, we employed a proprietary medical dataset consisting of 1.3 million real-world English medical audio recordings (all clinical audios were fully de-identified under HIPAA Safe Harbor; data collection followed
IRB-approved protocols) paired with human-annotated transcripts. These recordings were collected over a six-month period from more than 100 clinical offices, capturing a wide variety of speaker accents. The audio samples range from 6 seconds to 11 minutes and span a wide variety of clinical content, including symptoms, procedures, and medication names. Because the recordings originate from real clinical environments, they often contain background noise such as medical equipment sounds and office chatter.
This medical corpus provides rich acoustic diversity, which enhances the model's robustness and generalizability for audio processing in healthcare applications.



\subsection{Custom Training}
The model was trained using the AdamW optimizer with gradient norm clipping \cite{Pascanu} and a linear learning rate schedule that decays to zero following a warmup period spanning the first 512 updates. Training was conducted with a batch size of 64 over 4 epochs.
During the initial 2 epochs, we froze both the audio encoder and the language model, training only the adaptation layer to stabilize modality alignment. Subsequently, the audio encoder was fine-tuned for 1 epoch, followed by fine-tuning the LLM decoder for an additional epoch. Fine-tuning of the language model was performed using Low-Rank Adaptation (LoRA) \cite{LoRA}, allowing efficient adaptation with reduced computational cost.



\subsection{Results}
 
The primary evaluation metric for the public datasets (FLEURS and LibriSpeech) was the Word Error Rate (WER) \cite{jiwer}, which measures transcription accuracy. Additionally, we assessed the model on our proprietary medical dataset using both WER and Entity Word Error Rate (EWER). The EWER metric specifically evaluates entity-level transcription accuracy, focusing on clinically relevant entities such as medication names. For example, if an entity consists of two words (e.g., \textit{avacincaptad pegol}), and only one word (e.g., \textit{pegol}) is correctly transcribed, the EWER for that entity would be $\frac{1}{2}$. 

We used the implementation from the Jiwer Python library \cite{jiwer} to compute WER. To ensure fair and consistent evaluation across all ASR systems, we applied the
standardized post-processing steps (see \ref{sec:postprocessing} for more details) used in \cite{Adedeji24} to both hypothesis and reference transcripts. 
We implemented a custom script to compute EWER and contextualized our results by comparing them with SOTA models from the literature that provide open-source implementations or code, including United-MedASR \cite{United-MedASR}, Qwen2-Audio \cite{Qwen2}, Whisper-Large-V2 \cite{Radford2022}, Whispering-Llama \cite{radhakrishnan2023whispering}. 
We extracted their reported performance directly from their respective publications or run their open sourced models or code. A summary of the comparison is provided in Table~\ref{tab:results}. Overall, our model attains performance better than the SOTA methods on standard benchmark datasets, while offering substantial improvements on medical-domain transcription tasks.

\begin{table}[h]
  \centering
  \begin{tabular}{lccc}
    \hline
    \textbf{data} & \textbf{model}& \small{\textbf{WER}}& \small{\textbf{EWER}}\\
    \hline    
  LibriSpeech  &Whisper & 2.70 & NA    \\
  LibriSpeech  &United-MedASR & 0.98 & NA  \\
  LibriSpeech  & Qwen2-Audio & 1.61 & NA   \\
  LibriSpeech & W-Llama & 1.72 & NA \\
  LibriSpeech  & Au-M-ol & \textbf{0.42} & NA \\
  FLEURS  &Whisper & 5.64 & NA    \\
  FLEURS  &United-MedASR & 0.34 & NA  \\
  FLEURS  & Qwen2-Audio & 2.78  & NA   \\
  FLEURS & W-Llama & 2.26 & NA \\
  FLEURS  & Au-M-ol & \textbf{0.18} & NA \\
  Medical & Whisper & 3.07 & $40.79$ \\
  Medical  &United-MedASR & $2.85$ & $38.94$  \\
 Medical  & Qwen2-Audio &$3.01$ & $40.53$ \\ 
 Medical  & W-Llama &$3.28$ & $41.29$ \\ 
 Medical & Au-M-ol & \textbf{1.26} & \textbf{15.23} \\
   \hline
  \end{tabular} 
  \caption{Experimental results of Automatic Speech Recognition (ASR) across various datasets using Au-M-ol, compared with other state-of-the-art models. W-Llama is short for Whispring-Llama.}
  \label{tab:results}
\end{table}

We also benchmarked inference speed across models. All experiments were conducted on a cluster equipped with eight NVIDIA H100 GPUs (80 GB each). Table \ref{tab:time} reports the average inference time for transcribing one minute of medical audio. The results indicate that Au-M-ol achieves efficiency comparable to SOTA models, while maintaining strong transcription accuracy.

\begin{table}[ht]
  \centering
  \begin{tabular}{lc}
    \hline
    \textbf{model}& \textbf{time (s)} \\
    \hline
     Whisper &  12.1   \\
     United-MedASR & 11.8  \\
     Qwen2-Audio &  15.9  \\
     WhisperingLlama & 14.2  \\
     Au-M-ol & 12.7  \\ 
   \hline
  \end{tabular} 
  \caption{Average time (seconds) consumed for a 1 minute medical audio clip. }
  \label{tab:time}
\end{table}

\subsection{Ablation Studies and Model Fine-Tuning}

We conducted ablation experiments to assess the contribution of each component and its potential varieties in our model. The configurations and corresponding results are summarized in Table~\ref{tab:ablation}. In addition, we fine-tuned several state-of-the-art open-source models on the training split of our medical dataset and evaluated them on the test split. The outcomes are reported in Table~\ref{tab:finetune}. Across all comparisons, our proposed model consistently achieves the strongest overall performance among evaluated models on the medical ASR task.

\section{Conclusions}
\label{sec:conclusions}

We present Au-M-ol, a unified multimodal architecture that integrates speech and text processing for clinical applications. Unlike traditional ASR-LLM pipelines that rely on separate components for transcription and semantic understanding, our model fuses a Whisper-based audio encoder, a custom adaptation layer, and a decoder-only language model (e.g., LLaMA-3.1-8B) into a single, end-to-end system. This integration enables simultaneous processing of acoustic and textual inputs, supporting tasks such as automatic speech transcription
with reduced computational overhead and latency.
The adaptation layer aligns high-dimensional audio features with the LLM’s embedding space using dimensionality reduction and cross-modal attention. Both the audio and text representations are jointly optimized using an alignment loss and an output loss, facilitating robust multimodal understanding. 
By streamlining multimodal processing, Au-M-ol addresses key challenges in healthcare AI—including latency, scalability, and system complexity—while enabling real-time, context-aware decision support.

\section{Future Work}
\label{sec:futurework}

While Au-M-ol demonstrates strong performance in both general and medical-domain ASR tasks, several directions remain for future exploration. 

\textbf{Multilingual and multidialect extension}: Extending the model to support additional languages and dialects, especially underrepresented ones in clinical contexts would improve accessibility and generalizability across global healthcare settings. 

\textbf{Joint training with structured knowledge:} Integrating structured medical knowledge, such as UMLS \cite{Bodenreider2004} or drug ontologies \cite{Nelson2009}, may enhance entity recognition and support more accurate and informed generation in downstream tasks like summarization and question answering.

\textbf{Improved interpretability}: Developing methods for visualizing and interpreting cross-modal attention and decision pathways would increase clinical trust and help validate the model’s predictions.

\textbf{End-to-end clinical applications}: We aim to integrate Au-M-ol into real-time clinical workflows, such as diagnostic assistance, ambient documentation, or audio-based retrieval systems. This will require further validation in live clinical settings and attention to privacy, safety, and user experience.

\section{Limitations}
\label{sec:limitations}
Despite its advantages and promising performance, the proposed unified model has several limitations that warrant consideration:
\begin{itemize}
    \item Data requirements:
Training a joint audio-text model with meaningful alignment requires large amounts of high-quality, multimodal clinical data, which can be scarce or difficult to obtain due to privacy concerns and labeling costs. Limited data may restrict the model’s ability to generalize to diverse patient populations or rare conditions.
 \item Modality imbalance: Audio and text modalities have inherently different characteristics and information densities. The model may struggle when one modality is noisy, incomplete, or inconsistent with the other, potentially degrading performance if robust handling of missing or corrupted modalities is not explicitly addressed.
\item Interpretability and explainability: While the integrated architecture offers efficiency, it can obscure the individual contributions of audio and text features, complicating clinical interpretability. Understanding how the model derives specific predictions or summaries remains a challenge, which is critical in high-stakes healthcare environments.
\end{itemize}

\bibliography{custom}

@article{Karner,
  title="Towards Improving ASR Outputs of Spontaneous Speech with LLMs",
author={Manuel Karner and Julian Linke and Mark Kröll and Barbara Schuppler and Bernhard C. Geiger},
  journal={Proceedings of the 20th Conference on Natural Language Processing (KONVENS)},
  year={2024}
}

@article{Naderi2024,
  title="Towards interfacing large language models with ASR systems using confidence
measures and prompting",
author={Maryam Naderi and Enno Hermann and Alexandre Nanchen and Sevada Hovsepyan and  Mathew Magimai.-Doss},
  journal={InterSpeech},
  year={2024}
}

@article{mslam2022,
  title="mSLAM: Massively Multilingual Joint Pre-Training for Speech and Text",
author={Ankur Bapna and Colin Cherry and Yu Zhang and Ye Jia and Melvin Johnson and Yong Cheng and Simran Khanuja and Jason Riesa and Alexis Conneau},
  journal={arXiv preprint arXiv:2202.01374},
  year={2022}
}

@article{speechut2022,
  title={SpeechUT: Bridging Speech and Text with Hidden-Unit for Encoder-Decoder Based Speech-Text Pretraining},
  author={Shuai Zhang and Long Zhou and Junyi Ao and Shujie Liu and Lirong Dai and Jinyu Li and Furu Wei},
  journal={arXiv preprint arXiv:2210.03730},
  year={2022}
}

@article{speechllama2023,
  title={On decoder-only architecture for speech-to-text and large language model integration},
  author={Jian Wu and Yashesh Gaur and Zhuo Chen and Long Zhou and Yimeng Zhu and Tianrui Wang and Jinyu Li and Shujie Liu and Bo Ren and Linquan Liu and Yu Wu},
  journal={arXiv preprint arXiv:2307.03917},
  year={2023}
}

@article{ssr2024,
  title={SSR: Speech-Language Pretraining with Alignment-aware Modality Connector},
  author={Wei, Hao and et al.},
  journal={arXiv preprint arXiv:2410.00168},
  year={2024}
}

@article{dualdecoder2020,
  title={A Dual-decoder Transformer for Joint Automatic Speech Recognition and Robust Speech Translation},
  author={Zhang, Chengyi and et al.},
  journal={arXiv preprint arXiv:2011.00747},
  year={2020}
}

@article{cascaded2024,
  title={Cascaded Cross-Modal Transformer Network for Audio-Text Emotion Recognition},
  author={Shah, Alpa and Verma, Nishant and Joshi, Yash},
  journal={Artificial Intelligence Review},
  year={2024},
  publisher={Springer},
  doi={10.1007/s10462-024-10869-1}
}

@article{perceiver2021,
  title={Perceiver: General Perception with Iterative Attention},
  author={Jaegle, Andrew and et al.},
  journal={arXiv preprint arXiv:2103.03206},
  year={2021}
}

@article{United-MedASR,
  title={HIGH-PRECISION MEDICAL SPEECH RECOGNITION THROUGH
SYNTHETIC DATA AND SEMANTIC CORRECTION:
UNITED-MEDASR},
  author={Sourav Banerjee and Ayushi Agarwal and Promila Ghosh},
  journal={arXiv preprint arXiv:2412.00055},
  year={2024}
}

@misc{jiwer,
  author = {Jitsi},
  title = {jiwer: Evaluate your speech-to-text system with similarity measures such as WER},
  year = {2024},
  howpublished = {\url{https://github.com/jitsi/jiwer}}
}

@article{FLEURS,
  title={FLEURS: Few-shot Learning Evaluation of Universal Representations of Speech},
  author={Alexis Conneau and et al},
  journal={arXiv preprint arXiv:2205.12446},
  year={2022}
}

@article{Bodenreider2004,
  author  = {Olivier Bodenreider},
  title   = {The Unified Medical Language System ({UMLS}): integrating biomedical terminology},
  journal = {Nucleic Acids Research},
  year    = {2004},
  volume  = {32},
  number  = {Database issue},
  pages   = {D267--D270},
}

@article{Nelson2009,
  author  = {S. J. Nelson and N. A. Abraham},
  title   = {RxNorm: a normalized naming system for generic and branded drugs},
  journal = {Journal of the American Medical Informatics Association},
  year    = {2009},
  volume  = {16},
  number  = {3},
  pages   = {347--356},
}

@inproceedings{radhakrishnan2023whispering,
  title={Whispering LLaMA: A Cross-Modal Generative Error Correction Framework for Speech Recognition},
  author = {Srijith Radhakrishnan and Chao-Han Huck Yang and Sumeer Ahmad Khan and Rohit Kumar and Narsis A. Kiani and David Gomez-Cabrero and Jesper N. Tegner},
  booktitle={Proc. of EMNLP},
  year={2023}
}

@article{LoRA,
  title={LoRA: Low-Rank Adaptation of Large Language Models},
  author={Edward J. Hu and Yelong Shen and Phillip Wallis and Zeyuan Allen-Zhu and Yuanzhi Li and Shean Wang and Lu Wang and Weizhu Chen},
  journal={arXiv:2106.09685},
  year={2021}
}

@article{Adedeji24,
  title={The Sound of Healthcare: Improving Medical Transcription ASR Accuracy with Large Language Models},
  author={Ayo Adedeji and Sarita Joshi and Brendan Doohan},
  journal={arXiv:2402.07658},
  year={2024}
}

@article{lyu2023macaw,
  title={Macaw-LLM: Multi-Modal Language Modeling with Image, Audio, Video, and Text Integration},
  author={Lyu, Chenyang and Wu, Minghao and Wang, Longyue and Huang, Xinting and Liu, Bingshuai and Du, Zefeng and Shi, Shuming and Tu, Zhaopeng},
  journal={arXiv preprint arXiv:2306.09093},
  year={2023}
}

@article{liu2013,
    author  = "Kefei Liu and João Paulo C. L. da Costa and Hing Cheung So and ‪Andre L. F. de Almeida",
    title   = "Semi-blind receivers for joint symbol and channel estimation in space-time-frequency MIMO-OFDM systems",
    year    = "2013",
    journal = "IEEE Transactions on Signal Processing"
}

@article{liu2016,
    author  = "Kefei Liu and Joao Paulo C. L. da Costa and Hing Cheung So and Lei Huang and Jieping Ye",
    title   = "Detection of number of components in CANDECOMP/PARAFAC models via minimum description length",
    year    = "2016",
    journal = "Digital Signal Processing"
}

@article{Du20,
    author  = "Lei Du and Kefei Liu and Xiaohui Yao and Shannon L. Risacher and Junwei Han and Andrew J. Saykin and Lei Guo and Li Shen",
    title   = "Detecting genetic associations with brain imaging phenotypes in Alzheimer’s disease via a novel structured SCCA approach",
    year    = "2020",
    journal = "Medical image analysis"
}

@inproceedings{yang2024malaasr,
      title={MaLa-ASR: Multimedia-Assisted LLM-Based ASR}, 
      author={Guanrou Yang and Ziyang Ma and Fan Yu and Zhifu Gao and Shiliang Zhang and Xie Chen},
      booktitle={Proc. INTERSPEECH},
      year={2024},
}

@inproceedings{Librispeech,
  title={Librispeech: An ASR corpus based on
public domain audio books},
  booktitle={IEEE International Conference on Acoustics, Speech and Signal Processing},
  year=2015,
  author={Vassil Panayotov and Guoguo Chen and Daniel Povey and Sanjeev Khudanpur}
}

@inproceedings{Pascanu,
  author = {Razvan Pascanu and Tomas Mikolov and Yoshua Bengio},
  title = {On the difficulty of training recurrent neural networks},
  booktitle = {Proceedings of the 30th International Conference on Machine Learning},
  pages = {1310--1318},
  year = 2012
}

@inproceedings{Smyrnakis2021,
  title={Primary care professionals’ experiences during the first wave of the COVID-19 pandemic in Greece: A qualitative study},
  author={ E. Smyrnakis and et al.},
  booktitle={BMC Family Practice},
  pages={1--10},
  year={2021},
}

@inproceedings{Zhang2020,
  title={Physician workforce in the United States of America: Forecasting nationwide shortages},
  author={Xiaoming Zhang and Daniel Lin and Hugh Pforsich and Vernon W. Lin},
  booktitle={Human Resources for Health}, 
  year={2020},
}

@inproceedings{Radford2022,
  title={Robust Speech Recognition via Large-Scale Weak Supervision},
  author={Alec Radford and Jong Wook Kim and Tao Xu and Greg Brockman and Christine McLeavey and Ilya Sutskever},
  booktitle={Audio and Speech Processing}, 
  year={2022},
}

@inproceedings{Google2022,
  title={MedPalm: A language model for healthcare},
  author={Google},
  booktitle={Google Research Blog}, 
  year={2022},
}

@inproceedings{Qwen2,
  title={Qwen2-Audio Technical Report},
  author={Yunfei Chu and Jin Xu and Qian Yang and Haojie Wei and
Xipin Wei and Zhifang Guo and Yichong Leng and Yuanjun Lv and Jinzheng He and
Junyang Lin and Chang Zhou and Jingren Zhou},
  booktitle={arxiv:2407.10759}, 
  year={2024},
}

@inproceedings{Chu2023,
  title={Qwen-Audio: Advancing Universal AudioUnderstanding
via UnifiedLarge-Scale Audio-LanguageModels},
  author={Yunfei Chu and Jin Xu and Xiaohuan Zhou and Qian Yang and
Shiliang Zhang and Zhijie Yan and Chang Zhou and JingrenZhou},
  booktitle={arxiv:2311.07919v2}, 
  year={2023},
}

\appendix

\section{Appendix}

\subsection{Post-processing for transcriptions}
\label{sec:postprocessing}
To ensure fair and consistent evaluation across all ASR systems, we applied the following
standardized post-processing steps to both hypothesis and reference transcripts:

\begin{enumerate}
    \item \textbf{Disfluency Removal:} Fillers such as ``ummm'', and ``ahh'' were removed from all transcripts.
    \item \textbf{Numeral Normalization:} Numeric numbers (e.g., ``0'', ``16'') were converted into their written forms (``zero'', ``sixteen'').
    \item \textbf{Punctuation and Case Normalization:} Text was lowercased, spacing was standardized, and all punctuation was stripped.
    \item \textbf{Spelling Normalization:} British and American spelling variants were harmonized.
    \item \textbf{Hyphenation Normalization:} Hyphenated words were separated into two tokens to account for differences in hyphenation handling across ASR systems.
\end{enumerate}

\subsection{Ablation studies on model components}
\label{subsec:abl}
 
We conducted a series of ablation studies to assess how different architectural and component choices affect the final model performance. The major evaluated configurations are listed below, and the corresponding results are summarized in Table~\ref{tab:ablation}. 

\begin{itemize}
\item \textbf{No. 1}: Remove one fully connected layer from the adaptation layer. 
\item \textbf{No. 2}: Remove the Relu activation function in the adaptation layer. 
\item  \textbf{No. 3}: Remove the alignment loss term by setting $\alpha = 0$.
\item \textbf{No. 4}: Increase the weight of the alignment loss to $\alpha = 2$.
\item \textbf{No. 5}: Replace the Whisper-Large-V2 encoder with alternative Whisper encoder versions (medium).
\item \textbf{No. 6}: Replace the LLaMA-3.1-8B decoder with LLaMA-3-8B (we also tried using Mistral models and found the performance is generally worse compared with using Llama models).
\end{itemize} 

\begin{table}
  \centering
  \begin{tabular}{lccc}
    \hline
    \textbf{data} & \textbf{ablation}&\textbf{WER}& \textbf{EWER}\\
    \hline 
  LibriSpeech  &No. 1 & 1.27 & NA    \\
  LibriSpeech  &No. 2& 1.59 & NA  \\
  LibriSpeech  &No. 3& 1.03 & NA  \\
  LibriSpeech  &No. 4 & 0.69 & NA   \\
  LibriSpeech & No. 5 & 0.97 & NA \\
  LibriSpeech  & No. 6 & 0.51 & NA \\
  FLEURS  &No. 1 & 0.75 & NA    \\
  FLEURS  &No. 2 & 1.37 & NA    \\
  FLEURS  &No. 3 & 0.64 & NA  \\
  FLEURS  &No. 4 & 0.32  & NA   \\
  FLEURS & No. 5 & 0.24 & NA \\
  FLEURS  &No. 6 & 0.21 & NA \\
  Medical & No. 1 & 2.57 & $31.98$ \\
  Medical & No. 2 & 3.39 & $37.74$ \\
  Medical  &No. 3 & $2.33$ & $28.46$  \\
  Medical  &No. 4&$1.56$ & $17.49$ \\ 
  Medical  &No. 5 &$1.32$ & $16.65$ \\ 
  Medical & No. 6 & 1.29 & 15.38 \\
   \hline
  \end{tabular} 
  \caption{Ablation study results. Each ablation corresponds to the modifications described in the \ref{subsec:abl}.}
  \label{tab:ablation}
\end{table}

We also fine-tuned several state-of-the-art open-source models—Whisper-large-v2, Qwen2-Audio, and Whispering-LLaMA—on the training split of our medical dataset and evaluated their performance on the test split. The results are presented in Table~\ref{tab:finetune}. As shown in this Table, all models benefit substantially (compared with Table \ref{tab:results}) from fine-tuning, with notable reductions in both WER and EWER.  

\begin{table}[h]
\centering
\begin{tabular}{lcc}
\hline
\textbf{Model} & \textbf{WER} & \textbf{EWER} \\
\hline
Whisper-large-v2 & 1.58 & 18.72 \\
Qwen2-Audio & 1.64 & 16.39 \\
Whispering-LLaMA & 1.34 & 15.81 \\
Au-M-ol & 1.26 & 15.23 \\
\hline
\end{tabular}
\caption{Performance of fine-tuned SOTA models on the medical test split.}
\label{tab:finetune}
\end{table}

\end{document}